\documentclass[lettersize,journal]{IEEEtran}
\usepackage{amsmath,amsfonts}
\usepackage{algorithmic}
\usepackage{amssymb}
\usepackage{algorithm}
\usepackage{array}
\usepackage[caption=false,font=normalsize,labelfont=sf,textfont=sf]{subfig}
\usepackage{textcomp}
\usepackage{stfloats}
\usepackage{url}
\usepackage{verbatim}
\usepackage{graphicx}
\usepackage{cite}
\usepackage{tcolorbox}
\usepackage{ulem}

\begin{document}

\title{Detailed Object Description with Controllable Dimensions}

\author{Xinran~Wang,~Haiwen~Zhang,~Baoteng~Li,~Kongming~Liang,~Hao~Sun,~Zhongjiang~He,\\
~Zhanyu~Ma,~\IEEEmembership{Senior Member,~IEEE}, and Jun~Guo,~\IEEEmembership{Senior Member,~IEEE}
\thanks{X. Wang, H. Zhang, B. Li, K. Lang, Z. He, Z. Ma, J. Guo are with the Pattern Recognition and Intelligent
System Laboratory, School of Artificial Intelligence, Beijing University of Posts and Telecommunications, Beijing 100876, China. Z. He is also with China Telecom Artificial Intelligence Technology Co. Ltd, Beijing 100034, China. E-mail: \{wangxr, zhanghaiwen, meltry.lbt, liangkongming, hezhongjiang, mazhanyu, guojun\}@bupt.edu.cn}
\thanks{H. Sun is with the China Telecom Artificial Intelligence Technology Co. Ltd, Beijing 100034, China. E-mail: sunh10@chinatelecom.cn}%
\thanks{Corresponding author: Kongming Liang.}}

\markboth{Journal of \LaTeX\ Class Files,~Vol.~14, No.~8, August~2021}%
{Shell \MakeLowercase{\textit{et al.}}: A Sample Article Using IEEEtran.cls for IEEE Journals}


\maketitle

\begin{abstract}
Object description plays an important role for visually impaired individuals to understand and compare the differences between objects.
Recent multimodal large language models (MLLMs) exhibit powerful perceptual abilities and demonstrate impressive potential for generating object-centric descriptions.
However, the descriptions generated by such models may still usually contain a lot of content that is not relevant to the user intent or miss some important object dimension details. Under special scenarios, users may only need the details of certain dimensions of an object. 
In this paper, we propose a training-free object description refinement pipeline, \textbf{Dimension Tailor}, designed to enhance user-specified details in object descriptions. 
This pipeline includes three steps: dimension extracting, erasing, and supplementing, which decompose the description into user-specified dimensions. Dimension Tailor can not only improve the quality of object details but also offer flexibility in including or excluding specific dimensions based on user preferences. 
We conducted extensive experiments to demonstrate the effectiveness of Dimension Tailor on controllable object descriptions. Notably, the proposed pipeline can consistently improve the performance of the recent MLLMs.
The code is currently accessible at \url{https://github.com/xin-ran-w/ControllableObjectDescription}.
\end{abstract}


\section{Introduction}

\IEEEPARstart{D}{escribing} objects in images is an important and long-term research topic in computer vision~\cite{farhadi-2009-CVPR-describing}. With the description of objects, we can understand and compare the differences between objects, which is of great significance across various fields. For example, object description can help visually impaired individuals~\cite{Visually-Impaired} in perceiving the surroundings even when they are faced with a new kind of object. In terms of perception ability, an excellent object description model should be able to perceive details of various dimensions of the object, such as color, shape, and other important dimensions. In terms of human-computer interaction, the model should be more interactive to humans, that is, the description generation process is controllable by human intent.

\begin{figure*}[t!]
  \centering
\includegraphics[width=\linewidth]{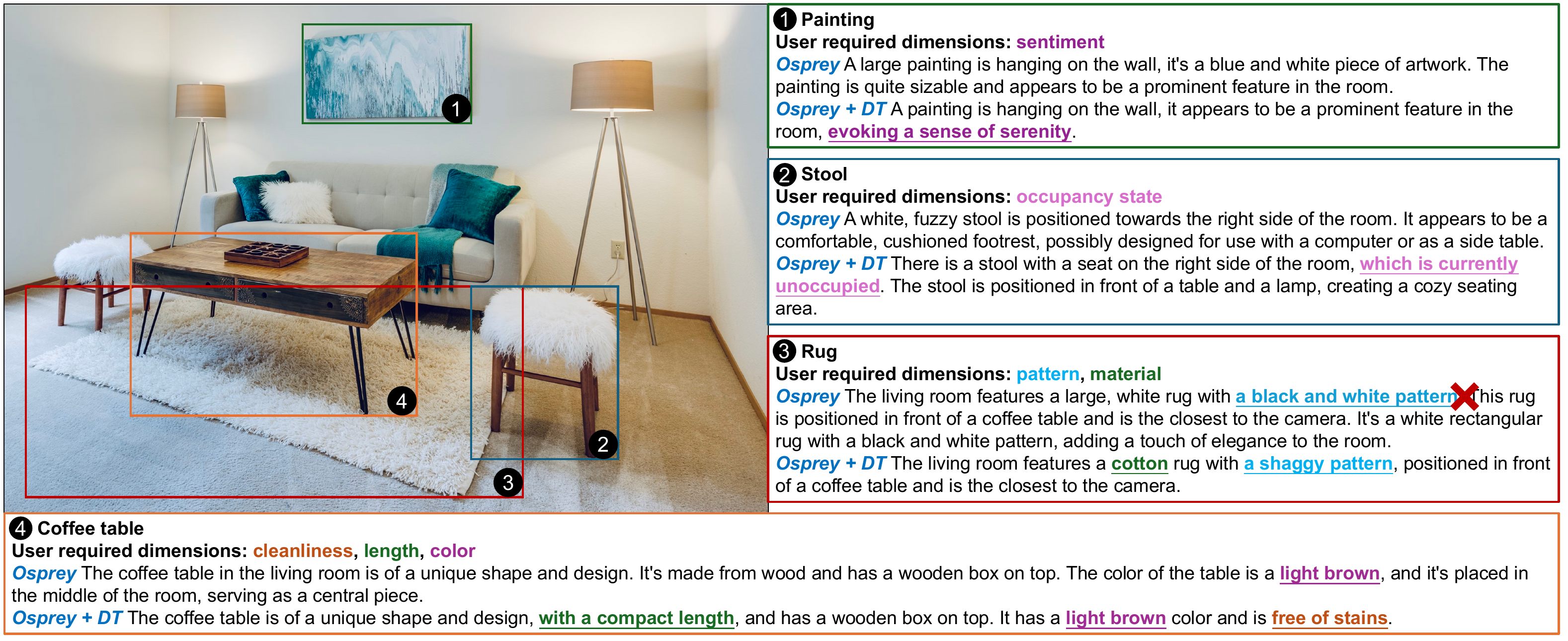}
  \caption{Dimensional controllable object description. In real-world scenarios, users need descriptions of objects focusing on specific dimensions of interest. However, existing multimodal large models often overlook dimension information user needs or include irrelevant information, resulting in descriptions that do not align with user preferences. By applying Dimension Tailor to MLLM-generated detailed object descriptions, the refined descriptions are more aligned with the user-specified dimension, reducing redundancy and focusing on the desired object dimensions.}
  \label{fig:abstract}
\end{figure*}

Recently, multimodal large language models (MLLMs) \cite{liu-2023-NeurIPS-LLaVA, dai-2023-arXiv-instructblip, bai-2023-arXiv-qwen, you-2023-arXiv-Ferret, yuan-2023-arXiv-Osprey, zhang-2023-arXiv-gpt4roi, openai-gpt4o, openai-gpt4} have shown outstanding performance in different visual tasks. They can provide users with convenient interactive experiences through language or spatial prompts. 
For example, MLLMs supporting pixel-level understanding, e.g. Osprey \cite{yuan-2023-arXiv-Osprey}, receive spatial cues from user input and can focus the described content within the object region. Despite significant advances, there are two main problems with these MLLMs:
\textbf{(1) Object descriptions generated by MLLMs may still ignore the dimensions requested by the user.} Taking the first instance (Painting) in Fig.\ref{fig:abstract} as an example, the \textit{sentiment} required by the user is not mentioned in the Osprey-generated description; 
\textbf{(2) The generated descriptions are usually redundant and fail to emphasize the user-specified object dimensions.} As shown in example 3 in Fig.\ref{fig:abstract}, users may only require details about specific semantic dimensions of an object in certain scenarios. However, Osprey generates irrelevant content, such as the \textit{color} of the rug, which is unrelated to the user's needs. This issue becomes particularly problematic when users attempt to locate specific information within hyper-detailed descriptions generated by MLLMs.

Different types of control signals are applied to represent user intent for object description, such as scene graphs~\cite{chen-2020-CVPR-sayaswish}, semantic roles~\cite{chen-2021-CVPR-human}, length \cite{deng-2020-length-cic, ding-2024-TPAMI-adaptive-length-cic}, sentiment \cite{mathews-2015-AAAI-senticap} and region prompt \cite{sun-2024-CVPR-alphaclip, zhang-2023-arXiv-gpt4roi, yuan-2023-arXiv-Osprey, you-2023-arXiv-Ferret, zhiliang-2023-arXiv-kosmos-2}.
However, such paradigms can hardly be adapted to the recent multimodal large language models as they require extensive extra training. 
One feasible way to customize MLLM to describe user intent is to incorporate relevant dimensions into language instructions \cite{liu-2023-NeurIPS-LLaVA}. However, due to the simple form of instructions used in the supervised fine-tuning (SFT) stage, MLLM-generated descriptions can hardly align with user intent. Therefore, MLLMs sometimes fail to follow users' instructions and tend to output either redundant or incomplete object descriptions.

In this paper, we seek a low-cost and effective way to enable users to control the MLLM-generated content. Since object descriptions are composed of various dimensions \cite{bravo-2023-CVPR-OVAD, chen-2023-CVPR-ovarnet, liang-2023-MM-HVAW, patterson-2016-ECCV-coco}, it is reasonable to take object dimensions as the control signal. To achieve this, we introduce \textbf{Dimension Tailor (DT)}, a training-free pipeline that can refine the MLLM-generated description to align with user intent. It enables users to control object description content by manipulating object dimension details.
DT has three main steps: dimension extracting, erasing, and supplementing. Dimension extracting aims to transform the long description into dimension tuples. Based on that, dimension erasing eliminates incorrect concepts or irrelevant dimensions. Dimension supplementing adds the dimensions not presented in the description but specified by users. In the end, the pipeline rewrites the description regarding the output dimensions.


In summary, our \textbf{main contributions} are: 
(1) We propose \textbf{Dimension Tailor, the first training-free description refinement pipeline} to enable users to control dimension details in object descriptions. Extensive experiments have verified the effectiveness of the proposed pipeline in enhancing recent MLLMs on controllable object descriptions.
(2) We propose \textbf{three evaluation metrics} to rigorously assess the controllability of MLLMs for object descriptions. We design \textbf{mean dimensional recall (mDR)} which measures the description covers all user-specified dimensions to reflect completeness; \textbf{mean dimensional precision (mDP)} which measures the description avoids irrelevant dimensions to reflect conciseness; \textbf{mean dimensional F1 score (mDF1)}, the harmonic mean of dimensional precision and recall, considering completeness and conciseness simultaneously to reflect the overall performance.
(3) 
We evaluate both open-source MLLMs and closed-source commercial MLLMs. Experimental results show that open-source MLLMs tend to overemphasize certain object-dimension compositions over others and lag behind commercial MLLMs. Notably, our proposed pipeline can improve open-source MLLMs controllability to the level of commercial MLLMs.

\section{Related Work}

\subsection{Controllable Image Captioning}

Image captioning aims to use language to express visual content in the image, which is the bridge between text modality and vision modality \cite{ke-2019-ICCV-reflective, cornia-2019-CVPR-show, chen-2021-CVPR-human, huang-2023-arXiv-ram++, zhang-2023-arXiv-ram, huang2023tag2text, sharma-2018-ACL-conceptual, schuhmann-2022-NeurIPS-laion5b, onoe-2024-arXiv-docci, doveh-2023-NIPS-DAC, bonilla-2023-arXiv-pixlore, anderson-2016-arXiv-spice, deng-2024-TMM-CDKM, liang-2024-TMM-IcoCap, wu-2021-TMM-fg-imgcap, yang-2024-TMM-cd-imgcap, li-2022-ICML-blip, garg-2024-EMNLP-imageinwords, zhao-TMM-MMKG-2024}. However, most image captioning models are intention-agnostic and cannot generate diverse descriptions according to different user intentions. Controllable image captioning is proposed to mimic caption diversity when humans describe images \cite{farhadi-2009-CVPR-describing, wang-2019-CVPR-describing-like-humans, chen-2020-CVPR-sayaswish}. Controllable image captioning enables users to actively control the image captioning process by representing their intent with control signals. There are many aspects of a caption that can be controlled \cite{bai-2023-arXiv-qwen, wen-2024-arXiv-complex-ins}, such as the content, structure, and semantics. Region captioning \cite{yuan-2023-arXiv-Osprey, Johnson-2016-CVPR-DenseCap, you-2023-arXiv-Ferret, zhang-2023-arXiv-gpt4roi, zhao-2024-arXiv-controlcap, zhang-2024-NIPS-omgllava, zhang-2024-CVPR-prompthighlighter} is a kind of CIC task aims to make the caption content focus on the region-of-interest in the image. Length-controllable image captioning \cite{deng-2020-length-cic, ding-2024-TPAMI-adaptive-length-cic, dwibedi-2024-NIPS-flexcap} can generate captions of different lengths for the same image. Sentiment controllable image captioning aims to mimic the different human subjective emotions in describing images \cite{mathews-2015-AAAI-senticap}. Caption Anything\cite{wang-2023-arXiv-caption-anything} combines multiple types of control signals, so users can control the sentiment, length, and content at the same time. Although region control can generate object-level captioning, it cannot control dimension-level details in object captions. In this paper, we explore ways to control the object description content by inputting user-specified dimensions.

\subsection{Detail Object Description Ability of Multimodal Large Language Models}

Describing objects with attributes \cite{gerych-2023-NeurIPS-debiasing, liang-2018-PAMI-unifying, wang-2023-NeurIPS-learning, yin-2024-CVPR-PAEdit, hu-2023-ICCV-tifa, patterson-2016-ECCV-coco, chen-2023-CVPR-ovarnet} is essential for comprehensively understanding the object. The recent rise of multimodal large language models has shown a surprising ability to generate long and detailed image captions. Its interactive instruction design allows users to control most aspects of the description in a natural language way. However, current open-source MLLMs show poor instruction following ability in following complex instructions \cite{qian-2024-arXiv-mia-bench, wen-2024-arXiv-complex-ins}.
Current MLLM benchmarks \cite{fu-2024-arXiv-MME, li-2023-arXiv-seed, li-2023-arXiv-seed2, li-2024-arXiv-seed2plus, yin-2023-arXiv-woodpecker, jing-2024-arXiv-faithscore, li-2023-arXiv-POPE, Cho-2024-ICLR-DSG, hu-2023-ICCV-tifa} and metrics mainly focus on visual question answering ability and hallucination in MLLM generated answers. And fewer benchmarks \cite{qian-2024-arXiv-mia-bench, mathews-2015-AAAI-senticap, wen-2024-arXiv-complex-ins} consider the instruction following ability. All the above benchmarks are not able to evaluate the dimensional controllability of MLLM in generating object descriptions. In this paper, we propose three evaluation metrics to measure the dimensional controllability.

\section{Controllable Object Description}
\label{sec:benchmark}

In this section, we first define the problem of controllable object description. Then, we introduce the proposed Dimension Tailor, a training-free description refinement pipeline, which contains three steps. Finally, we detailed how to evaluate the controllability and quality of MLLM-generated descriptions and our pipeline-refined descriptions. 

\subsection{Problem Formulation}
\label{subsec:formulation}

Given an image $x$, the target object label $o$ with location information $l$ (e,g., the object bounding box), and user intent $\mathcal{U}$, the task of controllable object description aims to generate a text sequence $d= \{w_1,w_2,...,w_T\}$ that accurately represents the visual content of $o$ within $x$ while adhering to the specified user intent $\mathcal{U}$. Here, we use user-specified dimensions as user intent $\mathcal{U}$. Finally, a large language model (LLM) is used to decode each word $w_t \in d$ based on previous words $w_{:<t}$ and other given inputs:

\begin{equation}
w_t=\operatorname{Decoder}\left(w_{:<t}, x, o, l, \mathcal{U}\right).
\label{eq:task-def-COD}
\end{equation}



\noindent{The} generated description $d$ should aim to maximize relevance to $\mathcal{U}$ while minimizing irrelevant or extraneous details. To achieve this, the generated description should be evaluated from the following aspects:

\begin{itemize}
    \item \textbf{Controllability}: The content of the generated description $d$ should align well with the user's intent. Controllability can be considered from two perspectives: (1) Completeness: Each specified dimension must be explicitly represented in $d$ to ensure the description accurately reflects the user's intent; (2) Conciseness: Details unrelated to $\mathcal{U}$ should be excluded to maintain focus and reduce redundancy.
    \item \textbf{Validity}: The content of $d$ should be free from hallucination and reflect the visual details of the object without any distortions or errors.
\end{itemize}

\noindent{In the context of multimodal large language models (MLLMs), we represent $\mathcal{U}$ in the form of a text prompt $p$ and use it to generate the description $d$.} For example, if the object $o$ is ``car" and $\mathcal{U}=\{\text{color}, \text{size}\}$, the input prompt will be ``Please describe the color and size of the car in detail.". 
In the next section, we introduce Dimension Tailor to refine the content of $d$ to align with the user intent.

\subsection{Dimension Tailor}
\label{sec:refinement}

\begin{figure*}[ht]
  \centering
  \includegraphics[width=\linewidth]{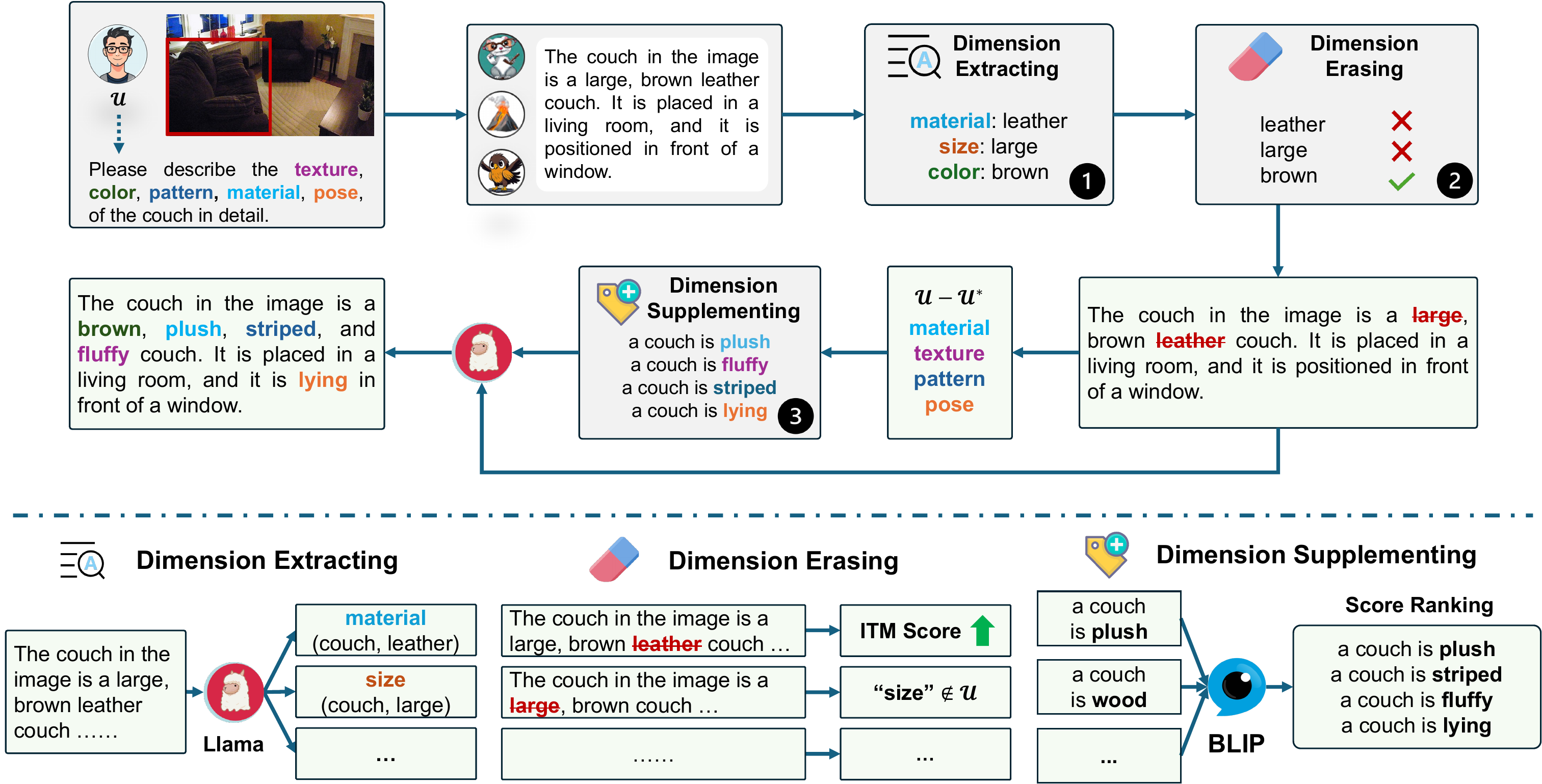}  
  \caption{The diagram of our description refinement pipeline, Dimension Tailor. The top half shows the full flow of the Dimension Tailor. The bottom half shows the detailed diagrams of the three key steps in Dimension Tailor. $\mathcal{U}$ is the user-required dimensions and $\mathcal{U}^*$ is the dimensions contained in MLLM generated description.}
  \label{fig:refine-method}
\end{figure*}

To enable MLLMs to generate object descriptions that align precisely with user intent, we present Dimension Tailor (DT), a training-free pipeline, to control the dimensional details in MLLM-generated object descriptions. The architecture of the proposed pipeline is illustrated in Fig.\ref{fig:refine-method}. Dimension Tailor allows for the dynamic addition and removal of content within object descriptions based on user intent. There are three steps in the proposed pipeline: dimension extracting, dimension erasing, and dimension supplementing. Each step plays a crucial role in controllable object description and will be introduced in the subsequent sections.

\textbf{Step 1: Dimension extracting.} The first step in refining an object description is to identify which dimensions are present. Inspired by \cite{Cho-2024-ICLR-DSG, yin-2023-arXiv-woodpecker, anderson-2016-arXiv-spice}, we utilize a dimension parser to parse the description into a list of structured tuples. Given the description $d$ and object $o$, the dimension parser function $G$ can be formulated as:
\begin{equation}
    G(o, d)=\left\{\left(o, u, a\right) \mid u \in \mathcal{U^*}, a \in \mathcal{A}_u\right\}.
\end{equation}
\noindent{where} $a$ is the attribute label of dimension $u$, $\mathcal{U^*}$ is the set of extracted dimensions from $d$ and $\mathcal{A}_u$ is the attribute label set of dimension $u$. 


As shown in the bottom half of Fig.\ref{fig:refine-method}, the output of dimension extracting is a list of tuples in the form of (object, dimension, attribute), e.g., (coach, material, leather) and (coach, size, large) can be extracted from the description ``There is a large leather couch in the image.''. 
After obtaining the dimension set $\mathcal{U^*}$ present in $d$, we can compare $\mathcal{U^*}$ with user actual intent dimension set $\mathcal{U}$ to guide subsequent refinement.

\textbf{Step 2: Dimension erasing.} There are two main problems in the description generated by MLLMs: (1) the output text is redundant and fails to align with the user-specified dimensions; (2) the output text contains inaccurate attributes even though they are relevant to user intent. Here, we introduce the erasure function to mitigate the above two problems. For the first problem, given the description $d$, and the output of dimension extracting step -- dimension set $\mathcal{U}^*$ and tuples set $G(o, d)=\left\{\left(o, u, a\right) \mid u \in \mathcal{U^*}, a \in \mathcal{A}_u\right\}$.
Compared to the user-specified dimension set $\mathcal{U}$, we can obtain the set of redundant dimension set by $\mathcal{U}^* \setminus \mathcal{U}$. For each tuple of redundant dimension, we prompt the LLM to erase the tuple-relevant content from the original description.
For the rest tuples of user-specified dimensions, we need to erase the tuple with the wrong attribute. Given a tuple $(o, u, a)$ extracted from $d$, we denote the description erasing this tuple as $d' = \operatorname{Erase}(d, u, a)$. We use BLIP's image-text matching (ITM) function to measure how well a description matches its corresponding image $x$. 
The ITM scores are used to determine if $d'$ matches the object-region $\operatorname{crop}(l, x)$ better than $d$, where $\operatorname{crop}(\cdot, \cdot)$ is the image crop function that extracts the object region in $x$. Formally, if $\operatorname{ITM}(d', \operatorname{crop}(l, x)) - \operatorname{ITM}(d, \operatorname{crop}(l, x)) > \tau_e $, where $\tau_e \in [0, 1]$ is the threshold of erasure, we can regard the tuple has a wrong attribute and erase the tuple, otherwise we keep it.

\textbf{Step 3: Dimension supplementing.} 
For the missing dimensions according to user intent, we propose a simple idea by leveraging BLIP as an attribute tagger. 
Due to the good performance of BLIP on attribute understanding \cite{bravo-2023-CVPR-OVAD}, we use its ITM function to tag the object with attributes of missing dimensions. 
Given the missing dimension set $u \in \mathcal{U} \setminus \mathcal{U^*}$, we use its attribute label set $\mathcal{A}_u$ and the object name to form phrases set $\mathcal{P}$, e.g., a car is black. Then we calculate the ITM score between the phrases and the cropped object region by $\operatorname{ITM}(p_a, \operatorname{crop}(l, x))$, where $p_a \in \mathcal{P}$ is the phase formed by attribute $a$. For each dimension $u$, we only add the attribute with the highest ITM score and the score must be larger than the supplement threshold $\tau_c$. Although setting supplement threshold $\tau_c$ may filter some noisy attributes, sometimes BLIP tags unreasonable attribute labels for the object. For example, the material dimension of a ``cup'' may be labeled with ``cloth''. To further improve the quality of supplemented attributes, we prompt the LLM to narrow the range of candidate attributes of dimension $t$ according to the object name. Due to the common sense knowledge in LLM, this operation can avoid adding some unreasonable attributes before doing attributes supplement. After the above three steps, we prompt the LLM to seamlessly integrate the supplemented attributes into the previously erased description, resulting in the final output. The whole algorithm is shown as algorithm \ref{algorithm:DT}.

\begin{algorithm}[t]
\caption{Dimension Tailor}
\label{algorithm:DT}
\begin{algorithmic}[1]
\REQUIRE Image $x$, Object label $o$, description $d$, user intent $\mathcal{U}$, object location $l$
\ENSURE Refined description $d_{\text{final}}$

\STATE \textbf{Step 1: Dimension Extracting}
\STATE Parse description $d$ using dimension parser $G$:
\[
G(o, d) = \left\{\left(o, u, a\right) \mid u \in \mathcal{U}^*, a \in \mathcal{A}_u\right\}
\]
\STATE Extract dimension set $\mathcal{U}^*$ from $d$
\STATE Compare $\mathcal{U}^*$ with user-specified dimensions $\mathcal{U}$ to guide refinement

\STATE \textbf{Step 2: Dimension Erasing}
\STATE Identify redundant dimensions: $\mathcal{U}^* \setminus \mathcal{U}$
\FOR{each redundant dimension $u \in \mathcal{U}^* \setminus \mathcal{U}$}
    \STATE $d = \operatorname{Erase} (d, u, a)$
\ENDFOR
\FOR{each tuple $(o, u, a) \in G(o, d)$ with $u \in \mathcal{U}^*$}
    \STATE Create a new description $d'$ by erasing $(o, u, a)$
    \IF{$\text{ITM}(d', \operatorname{crop}(l, x)) - \text{ITM}(d, \operatorname{crop}(l, x)) > \tau_e$}
        \STATE $d = d'$
    \ENDIF
\ENDFOR

\STATE \textbf{Step 3: Dimension Supplementing}
\STATE Identify missing dimensions: $\mathcal{U} \setminus \mathcal{U}^*$
\FOR{each missing dimension $u \in \mathcal{U} \setminus \mathcal{U}^*$}
    \STATE Generate attribute phrases using attribute label set $\mathcal{A}_{u}$
    \STATE Filter unreasonable candidate attributes
    \STATE Form the filtered attributes into phrase set $\mathcal{P}$
    \STATE Compute ITM scores for phrases and cropped image $x$
    \STATE Select the attribute $a$ with the highest ITM score if $\operatorname{ITM}(p_a, \operatorname{crop}(l, x)) > \tau_c$
\ENDFOR

\STATE Add all selected attribute $a$ to the description $d$ to obtain the refined description $d_{\text{final}}$
\RETURN Refined description $d_{\text{final}}$
\end{algorithmic}
\end{algorithm}

\subsection{Evaluation Metrics}
\label{sec:evaluation-metrics}
In this subsection, we detail the evaluation process for the generated descriptions. As outlined in the problem formulation (Section \ref{subsec:formulation}), the refined description must perform well in both two aspects: \textbf{controllability} and \textbf{validity}.
Controllability pertains to the description generation process's ability to effectively adhere to user-specified dimensions. On the other hand, correctness assesses the perception capability of the MLLM, ensuring that the attributes described accurately represent the visual content of the image.

\paragraph*{\textbf{Controllability}}

We introduce three metrics specifically designed to evaluate the dimension controllability of a set of object descriptions. Formally, given an object $o$, its description $d$, the potential dimension set $\mathcal{T}$ associated with $o$, the set of user-specified dimensions $\mathcal{U}$, and the set of dimensions contained in the description $\mathcal{U}^*$, we define the following for $t^\text{th}$ dimension $u_t \in \mathcal{T}$, where $ 0 < t \leq |\mathcal{T}|$:

\begin{itemize}

\item If $u_t \in \mathcal{U} \cap \mathcal{U}^*$, the description $i$ correctly includes dimension $t$, aligning with the user intent. This scenario represents a true positive sample for dimension $t$;

\item If $u_t \in \mathcal{U} \setminus \mathcal{U}^*$, the description $d$ fails to include the dimension $t$, indicating a false negative sample for dimension $t$.

\item If $u_t \in \mathcal{U}^* \setminus \mathcal{U}$, the inclusion of dimension $u$ in $d$ is unnecessary, thus representing a false positive for dimension $u_t$.

\end{itemize}

According to the above settings, for description set $\mathcal{D}$, we denote $\text{TP}_t$,  $\text{FP}_t$, and $\text{FN}_t$ as the number of true positive, false positive, and false negative descriptions for $t^\text{th}$ dimension $u_t$, respectively. \textbf{To reflect the completeness}, we design mean Dimensional Recall (mDR):
\begin{equation}
    \text{mDR} = \frac{1}{|\mathcal{T}|}\sum_{t=1}^{|\mathcal{T}|} \text{DR}_{u} = \frac{1}{|\mathcal{T}|}\sum_{t=1}^{|\mathcal{T}|}\frac{\text{TP}_t}{\text{TP}_t+\text{FN}_t},
\end{equation}
where $\text{DR}_{t}$ is the recall rate of Dimension $t$. It represents the probability dimension $t$ is described when it's in user intent. 
\textbf{To reflect the conciseness}, we design mean Dimensional Precision (mDP):
\begin{equation}
    \text{mDP} = \frac{1}{|\mathcal{T}|}\sum_{t=1}^{|\mathcal{T}|} \text{DP}_{t} = \frac{1}{|\mathcal{T}|}\sum_{t=1}^{|\mathcal{T}|}\frac{\text{TP}_t}{\text{TP}_t + \text{FP}_t},  
\end{equation}
where $\text{DP}_{t}$ indicates the likelihood that dimension $t$ is not redundant according to user intent.
Third, we use mean Dimensional F1 score \textbf{(mDF1)} to represent the overall performance:
\begin{equation}
    \text{mDF1} = \frac{1}{|\mathcal{T}|}\sum_{t=1}^{|\mathcal{T}|} \text{DF1}_{t} = \frac{1}{|\mathcal{T}|}\sum_{t=1}^{|\mathcal{T}|}\frac{2 \cdot \text{DP}_{t} \cdot \text{DR}_{t}}{\text{DP}_{t} + \text{DR}_{t}}.
\end{equation}
where the $\text{DF1}_{t}$ score is the harmonic mean of dimensional precision and recall of dimension $u_t$.

\paragraph*{\textbf{Validity}}

Evaluating the validity of a description involves verifying the correctness of each concepts it contains. Traditional evaluation metrics, such as BLEU \cite{papineni-2002-ACL-bleu}, METEOR \cite{banerjee-2005-ACL-meteor}, and CIDEr \cite{vedantam-2015-arXiv-cider}, which rely on n-gram overlap, have shown limitations in the context of detailed descriptions. These descriptions are often longer and contain more diverse vocabulary, making n-gram-based approaches less effective \cite{garg-2024-EMNLP-imageinwords, onoe-2024-arXiv-docci}. Model-based evaluation methods, such as CLIP-score \cite{hessel2021clipscore}, also face challenges, as the text encoder in CLIP has a limited context window (up to 77 tokens), making it unsuitable for processing lengthy descriptions.

To evaluate the correctness of the descriptions refined by Dimension Tailor, we employ a model-based evaluation framework inspired by LLaVA \cite{liu-2023-NeurIPS-LLaVA, liu-2023-arXiv-LLaVA-v1.5} and Vicuna \cite{chiang-vicuna-2023}, leveraging GPT-4 \cite{openai-gpt4} as the evaluator. The evaluation is structured as follows: 
\textbf{(1) Reference description generation}: GPT-4 synthesizes a detailed reference description of the target object based on expert-labeled attributes and object location information. This reference description serves as the evaluation benchmark, reflecting the ground-truth attributes; 
\textbf{(2) Description comparison}: Both the Dimension Tailor-refined description and the reference description are presented to GPT-4, along with the corresponding visual inputs, including attribute labels, image captions, and object bounding boxes; \textbf{(3) Scoring and analysis}: GPT-4 evaluates the correctness of the refined description by assigning a score on a scale of 1 to 10, with higher scores indicating better alignment with the reference description and expert annotations. Additionally, GPT-4 provides a detailed explanation of its evaluation to highlight the strengths and weaknesses of the refined descriptions. To compare with other methods based on the same reference description, we compute a relative score by dividing the score of the refined description by the score of the reference description.

By grounding the· evaluation in expert-labeled attributes and utilizing GPT-4's analytical capabilities, this framework ensures consistency across methods, as all evaluations are based on the same GPT-4-generated reference descriptions. This approach provides a robust and systematic measurement of the correctness of the refined object descriptions.

\begin{table}[t]
    \caption{The System Instruction For GPT-assisted Evaluation.}
    \small
    \begin{tcolorbox}
You are good at giving feedback on the performance of two AI assistants' generated detailed descriptions of a target region.
For your reference, the visual content in the image is represented with a few sentences describing the image.
In addition, objects around the target area and their coordinates are given. \textbf{And the most important reference: the object label and location and attributes of the target region are also given.}
All location coordinates are in the form of bounding boxes, represented as (x1, y1, x2, y2) with floating numbers ranging from 0 to 1. These values correspond to the top left x, top left y, bottom right x, and bottom right y. \\

\textbf{Please rate the responses of the assistants on a scale of 1 to 10}, where a higher score indicates better performance, according to the following criteria: \\

\textbf{Accuracy}: whether the response is accurate with respect to the image content. Responses with fewer hallucinations should be given higher scores. \\

Please output a single line for each criterion, containing only two values indicating the scores for Assistant 1 and 2, respectively. The two scores are separated by a space. Following the scores, please provide an explanation of your evaluation, avoiding any potential bias and ensuring that the order in which the responses were presented does not affect your judgment. \\

Output format:

Accuracy: [Score of Assistant 1] [Score of Assistant 2] \\
Reason: [Explanation of the evaluation]

    \end{tcolorbox}
    \label{eval-sys-msg}
\end{table}

\section{Experiments}

In this section, we introduce the evaluated MLLMs, followed by the implementation specifics of the Dimension Tailor and the dataset employed for evaluation. Our experiments begin with a thorough comparison of various MLLMs, supported by both quantitative metrics and qualitative assessments. Subsequently, we design experiments to showcase the enhancements achieved through the application of Dimension Tailor. Finally, we conduct ablation studies and provide visualizations to substantiate the effectiveness and robustness of our approach.

\subsection{Implementation Details} 
\paragraph*{\textbf{Details of MLLMs}} We evaluated six state-of-the-art MLLMs: LLaVAv1.5-7B \cite{liu-2023-NeurIPS-LLaVA, liu-2023-arXiv-LLaVA-v1.5}, GPT4ROI-7B \cite{zhang-2023-arXiv-gpt4roi}, AlphaCLIP + LLaVAv1.5-7B \cite{sun-2024-CVPR-alphaclip} (referred to as AlphaCLIP for brevity), Ferret-7B \cite{you-2023-arXiv-Ferret}, Osprey-7B \cite{yuan-2023-arXiv-Osprey}, and GPT-4o \cite{openai-gpt4o}. Among these models, all except LLaVAv1.5 and GPT-4o can process spatial prompts to focus on specific regions within an image. To maintain consistency, we used a box-cropped region as the image-level input for LLaVAv1.5 and GPT-4o. 

\paragraph*{\textbf{Details of Dimension Tailor}} The Dimension Tailor pipeline utilizes the ITM function of BLIP \cite{li-2022-ICML-blip} for erasing incorrect dimension tuples and supplementing new dimensions, and Llama3-8B \cite{LLama3-Github-llama3modelcard} for dimension extracting, dimension erasing, dimension filtering, and description rewriting. To assess controllability, we report three metrics outlined in Sec.~\ref{sec:evaluation-metrics}. 
Due to the high cost of the GPT-4 API, we conducted the GPT-assisted evaluation on 100 randomly selected samples from the 1k test set. The results presented in Table \ref{tab:refine-result} were obtained using three random seeds, with the mean and standard deviation reported. We also report $R_m$, the ratio of modified descriptions across all test samples.

\paragraph*{\textbf{Dataset}} We selected approximately 1,000 examples from the OVAD benchmark \cite{bravo-2023-CVPR-OVAD}, a fully annotated and high-quality attribute understanding benchmark created through crowdsourcing. The sample selection ensured a balanced distribution across different categories of object instances. 
The rationale behind this choice stems from two primary considerations: (1) OVAD covers 117 attribute classes on the 80 object classes of MSCOCO \cite{lin-2014-ECCV-COCO}, which is the best choice for evaluating the quality of attributes in object descriptions. (2) It offers a well-designed taxonomic framework for attributes, allowing us to directly utilize their dimensions as proxies for user intent.

\paragraph*{\textbf{Computation resources}}
All experiments were conducted using 2 NVIDIA Tesla A800 80G GPUs. However, the GPU memory required for a single task is less than 24G, making it possible to run on a single Nvidia GeForce RTX 3090 GPU.

\subsection{Comparison of Different MLLMs} 

\begin{figure}[t]
    \centering
    \includegraphics[width=\linewidth]{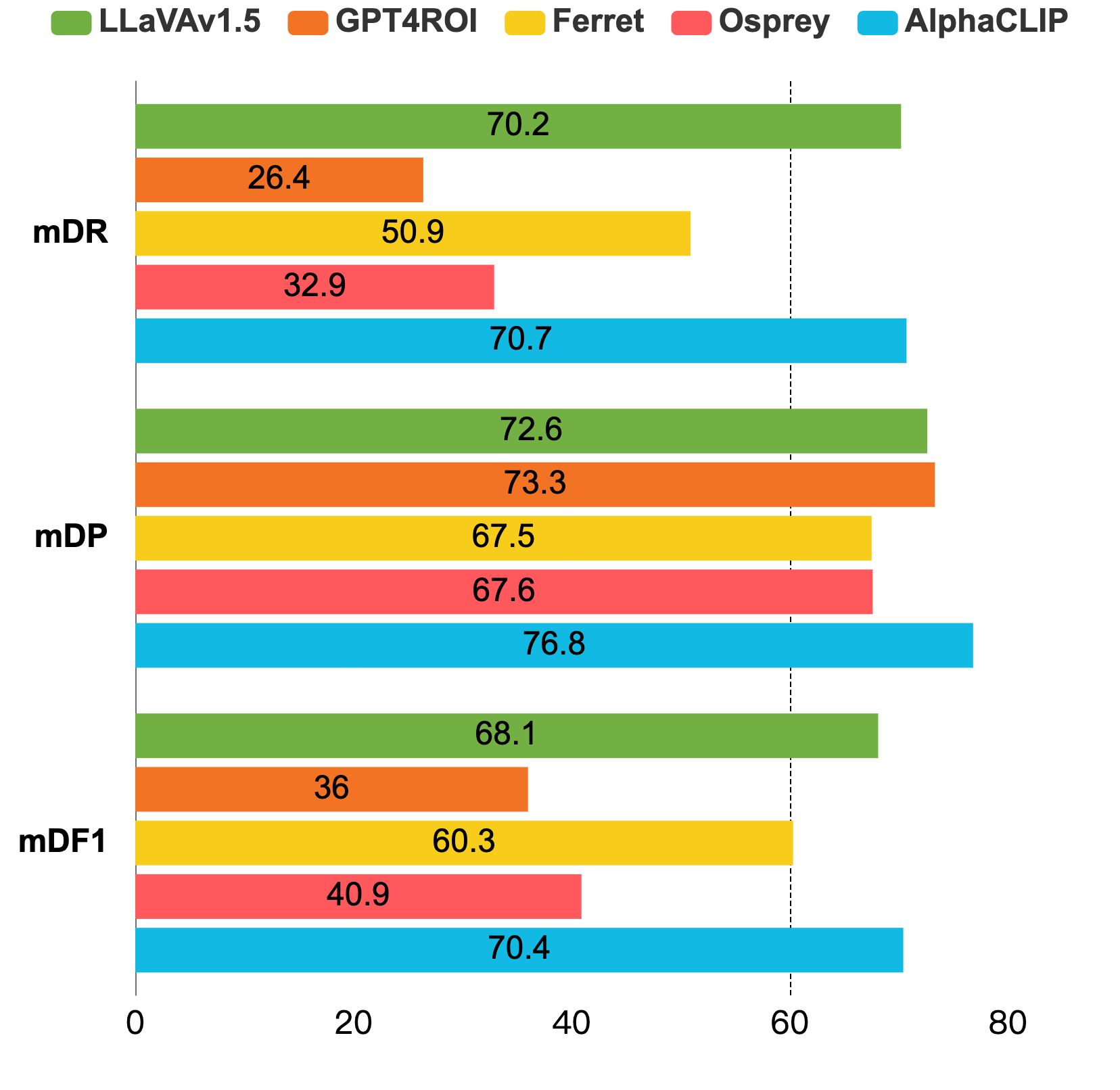}
    \caption{Controllability evaluation results of all open-source MLLMs.}
    \label{fig:ctrl-eval}
\end{figure}

\paragraph*{\textbf{Quantitative analysis}} As shown in Fig.\ref{fig:ctrl-eval}, indicated by the mDF1, AlphaCLIP + LLaVAv1.5-7B achieved the best overall performance on controllability among the open-source MLLMs. In contrast, GPT4ROI exhibits the lowest mDR value, suggesting that its generated object descriptions often lack many user intent dimensions. The higher mDP value for GPT4ROI can be attributed to its tendency to generate descriptions with fewer dimensions, thereby minimizing redundancy relative to user needs. We also evaluated the commercial closed-source model, OpenAI's GPT-4o, which is currently one of the most advanced MLLMs available. GPT-4o demonstrated exceptional performance in terms of controllability, achieving a high mDF1 score ($82.4 \pm 0.5$), significantly outperforming open-source MLLMs. These results underscore the dimensional controllability gap between open-source MLLMs and GPT-4o. Given that open-source 7B MLLMs are still limited in both the number of parameters and the complexity of their training instruction compared to GPT-4o, it is reasonable that GPT-4o has a much better performance on open benchmarks.

\begin{table}[t]
\centering
\caption{Controllability evaluation of refined object descriptions. DT is short for Dimension Tailor.}
\normalsize
\begin{tabular}{l|ccc}
    \hline \hline
    MLLMs & mDR & mDP & mDF1 \\ \hline

    \multicolumn{4}{l}{Open source} \\ \hline
    
    LLaVAv1.5 \cite{liu-2023-arXiv-LLaVA-v1.5} & $\mathbf{70.2 \pm 0.2}$ & $72.6 \pm 0.2$ & $68.1 \pm 0.2$         \\
    + DT       & $69.6 \pm 0.6$ & $\mathbf{85.0 \pm 0.2}$ & $\mathbf{74.2 \pm 0.3}$       \\ \hline
    
    GPT4ROI \cite{zhang-2023-arXiv-gpt4roi} & $26.4 \pm 0.3$ & $73.3 \pm 1.4$ & $36.0 \pm 0.5$           \\
    + DT       & $\mathbf{44.7 \pm 0.6}$ & $\mathbf{91.1 \pm 0.3}$ & $\mathbf{59.0 \pm 0.4}$          \\ \hline
    
    Ferret \cite{you-2023-arXiv-Ferret} & $50.9 \pm 0.5$ & $67.5 \pm 0.3$ & $60.3 \pm 0.2$   \\ 
    + DT      & $\mathbf{67.1 \pm 0.2}$ & $\mathbf{85.2 \pm 0.4}$  & $\mathbf{73.3 \pm 0.3}$   \\ \hline
    
    Osprey \cite{yuan-2023-arXiv-Osprey} & $32.9 \pm 0.3$ & $67.6 \pm 0.3$ & $40.9 \pm 0.3$   \\
    + DT       & $\mathbf{50.8 \pm 0.8}$ & $\mathbf{87.2 \pm 0.4}$ & $\mathbf{63.1 \pm 0.7}$    \\ \hline
    
    AlphaCLIP \cite{sun-2024-CVPR-alphaclip} & $70.7 \pm 1.0$ & $76.8 \pm 0.7$ & $70.4 \pm 0.9$  \\ 
    + DT       & $\mathbf{77.4 \pm 0.8}$ & $\mathbf{87.5 \pm 0.3}$ & $\mathbf{80.3 \pm 0.5}$             \\ \hline
    
    \multicolumn{4}{l}{Commercial} \\ \hline        
    GPT-4o \cite{openai-gpt4o} & $87.7 \pm 0.7$ & $79.3 \pm 0.6$ & $82.4 \pm 0.5$ \\
    
    \hline \hline
    \end{tabular} 
\label{tab:refine-result}
\end{table}

\begin{figure*}[htbp]
    \centering
    \includegraphics[width=\linewidth]{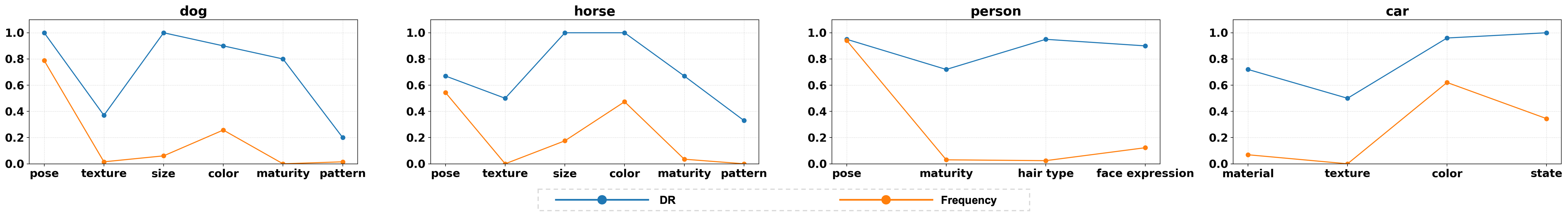}
    \caption{The DR and frequency in LLaVAv1.5 instruction tuning dataset of several common object-dimension combinations. The DR and frequency of different object-dimension combinations are positively correlated.}
    \label{fig:F-DR}
\end{figure*}

We further show the DR of several common object dimension combinations in  Fig.\ref{fig:F-DR}. The results reveal that MLLMs may exhibit inherent biases in their dimensional preferences when describing specific objects. For instance, AlphaCLIP tends to prioritize describing the \textit{pose} of a dog over its \textit{texture}. To gain deeper insights into this phenomenon, we analyzed the instruction-tuning dataset of LLaVAv1.5. By employing dimension extracting, we calculate the frequency of each dimension combination being described within the training instruction data. The similar trend of DR and frequency indicates that the dimensional biases present in the training data directly influence the MLLMs' preferences when describing specific objects.

\paragraph*{\textbf{Qualitative analysis}} In Fig.\ref{fig:desp-compare}, we further compare the descriptions of multiple MLLMs for the same object and user-specified dimensions. The visualized descriptions further validate that the GPT4ROI's response usually lacks some user-specified dimensions. For the dog in Fig.\ref{fig:desp-compare}(a), GPT4ROI only describes the three user-required dimensions. Compared to GPT4ROI and Ferret, AlphaCLIP's description is more in line with user intent. However, AlphaCLIP still ignores users' requests on some rarely described dimensions of the dog, such as, \textit{texture}.

\subsection{Performance of Dimension Tailor}

\begin{figure}[htbp]
    \centering
    \includegraphics[width=0.9\linewidth]{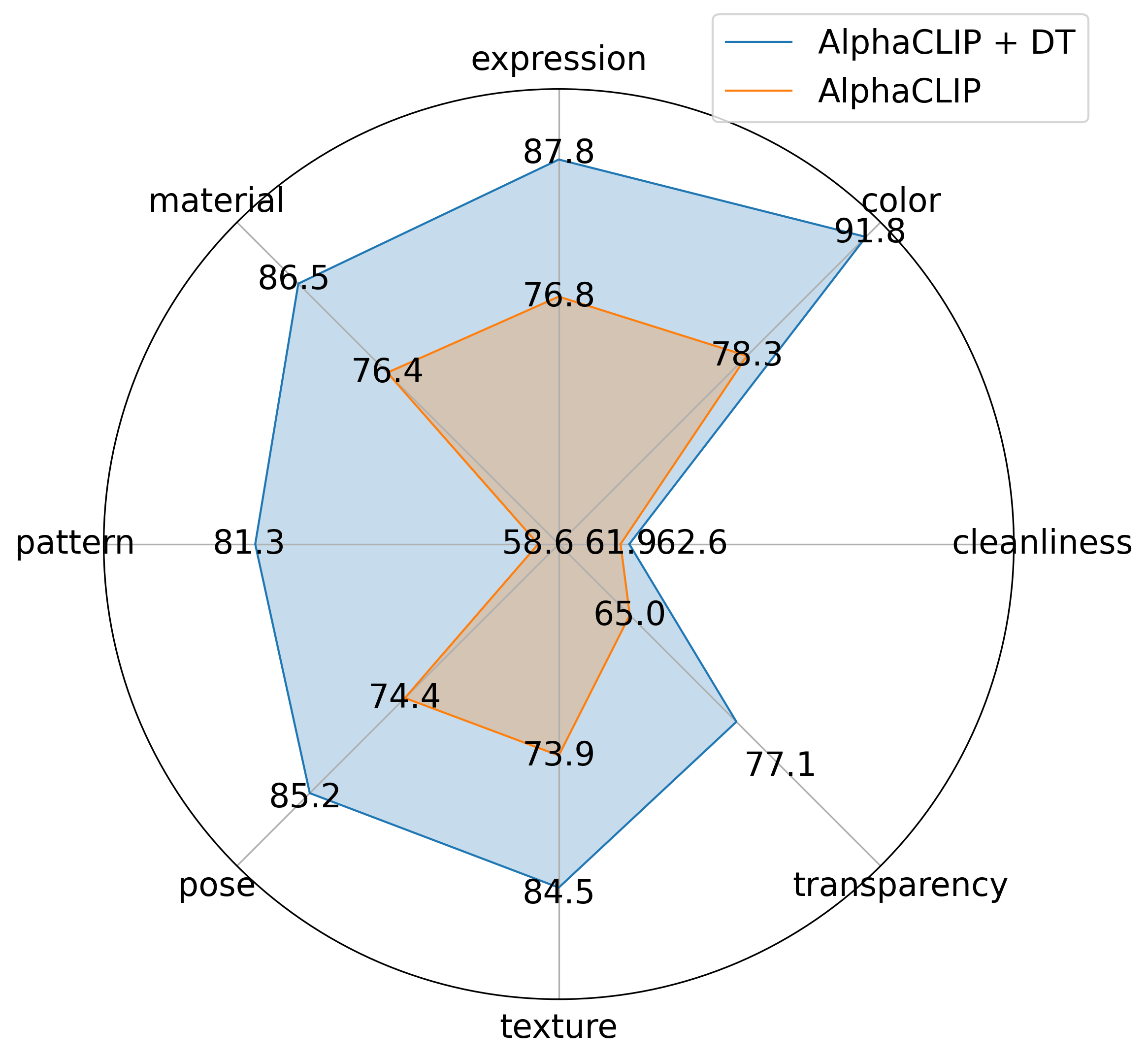}
    \caption{DF1 scores comparison of AlphaCLIP and AlphaCLIP + DT on various dimensions. }
    \label{fig:DF1-radar}
\end{figure}

\paragraph*{\textbf{Improvement analysis}} In Table \ref{tab:refine-result}, we present a comparative controllability analysis between the original and refined object descriptions generated by five MLLMs. The results demonstrate that our method significantly enhances controllability, as evidenced by the improved mDF1 scores across five open-sourced MLLMs. This improvement stems from our approach's ability to effectively eliminate redundant dimensions while supplementing attributes according to the user intent. Notably, our method enables smaller open-source models, such as AlphaCLIP-LLaVAv1.5-7B, to achieve a level of controllability that is comparable to that of the more advanced GPT-4o. To provide a more fine-grained view of some general dimensions' controllability, we compare the DF1 scores of original and refined descriptions of dimensions. As shown in Fig.\ref{fig:DF1-radar}, Dimension Tailor can effectively improve the controllability of each dimension.

\begin{figure}[htbp]
    \centering
    \includegraphics[width=\linewidth]{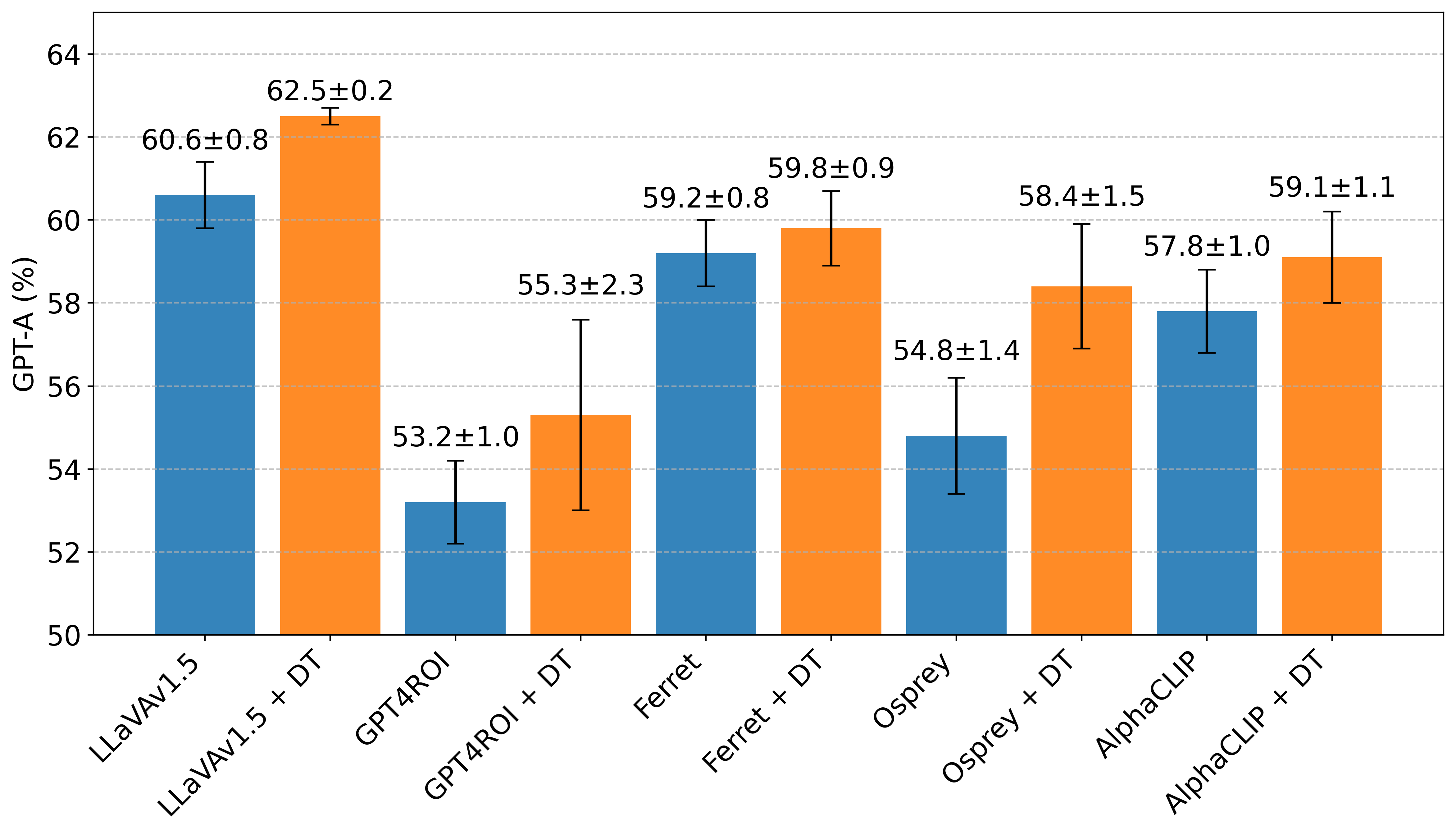}
    \caption{Quality evaluation of refined object descriptions. DT is short for Dimension Tailor.}
    \label{fig:GPT-A}
\end{figure}

The quality comparison is shown in Fig.\ref{fig:GPT-A}, the GPT-A results indicate that our method also contributes to a slight enhancement in the accuracy and relevance of attributes within the object descriptions generated by five MLLMs. Given the strong quality of controllability performance of AlphaCLIP, our subsequent analysis primarily focuses on this MLLM.

\begin{figure*}[htbp]
    \centering
    \includegraphics[width=\linewidth]{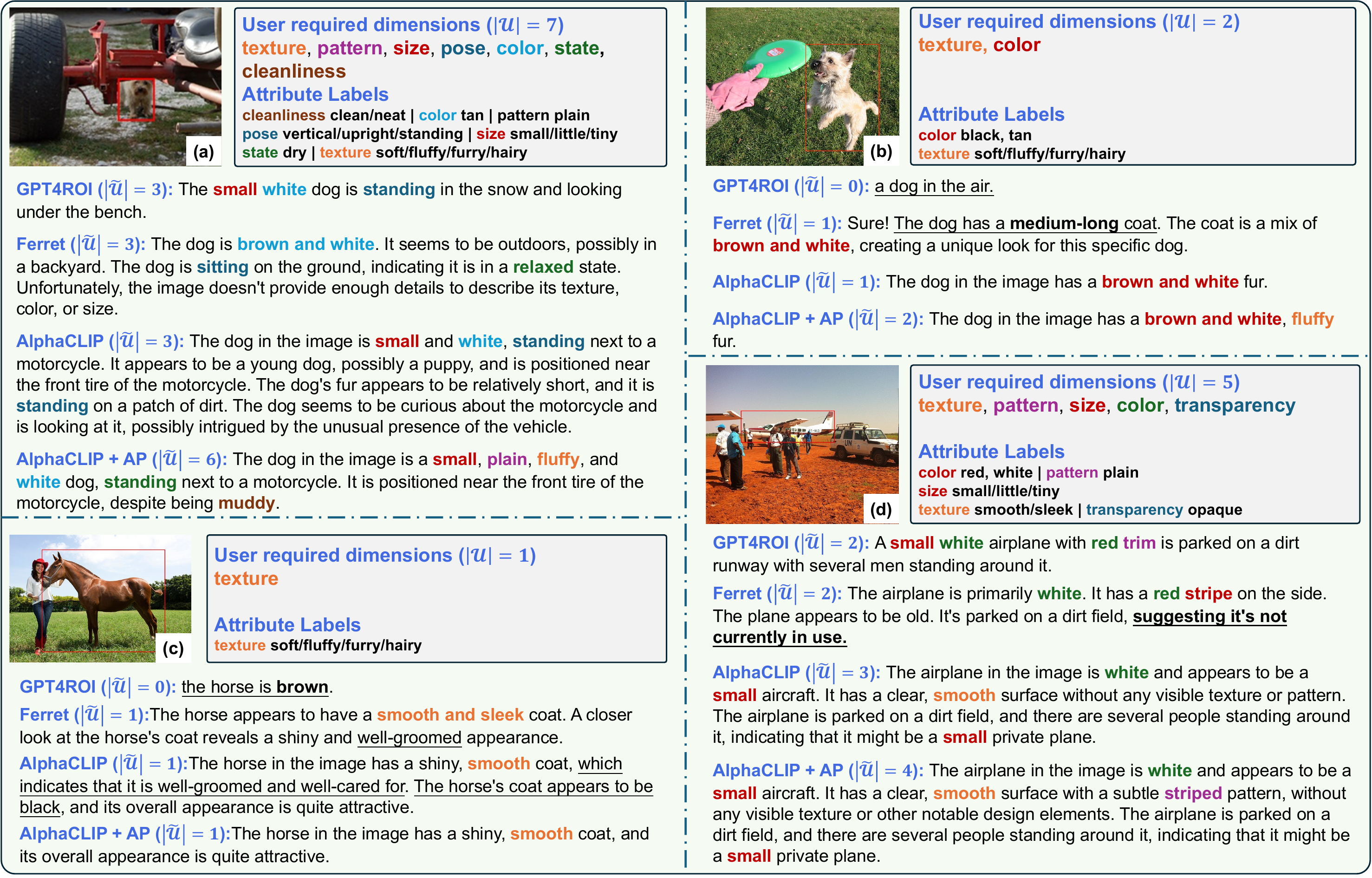}
    \caption{Visualization of controllable object descriptions generated by multiple MLLMs. $|\tilde{\mathcal{U}}|$ represent the cover number of user-specified dimensions, where $\tilde{\mathcal{U}}$ = $\mathcal{U}^* \cap  \mathcal{U}$. The underlined texts highlight unintended dimensions in the descriptions.}
    \label{fig:desp-compare}
\end{figure*}

\begin{figure*}[htbp]
  \centering
  \includegraphics[width=\linewidth]{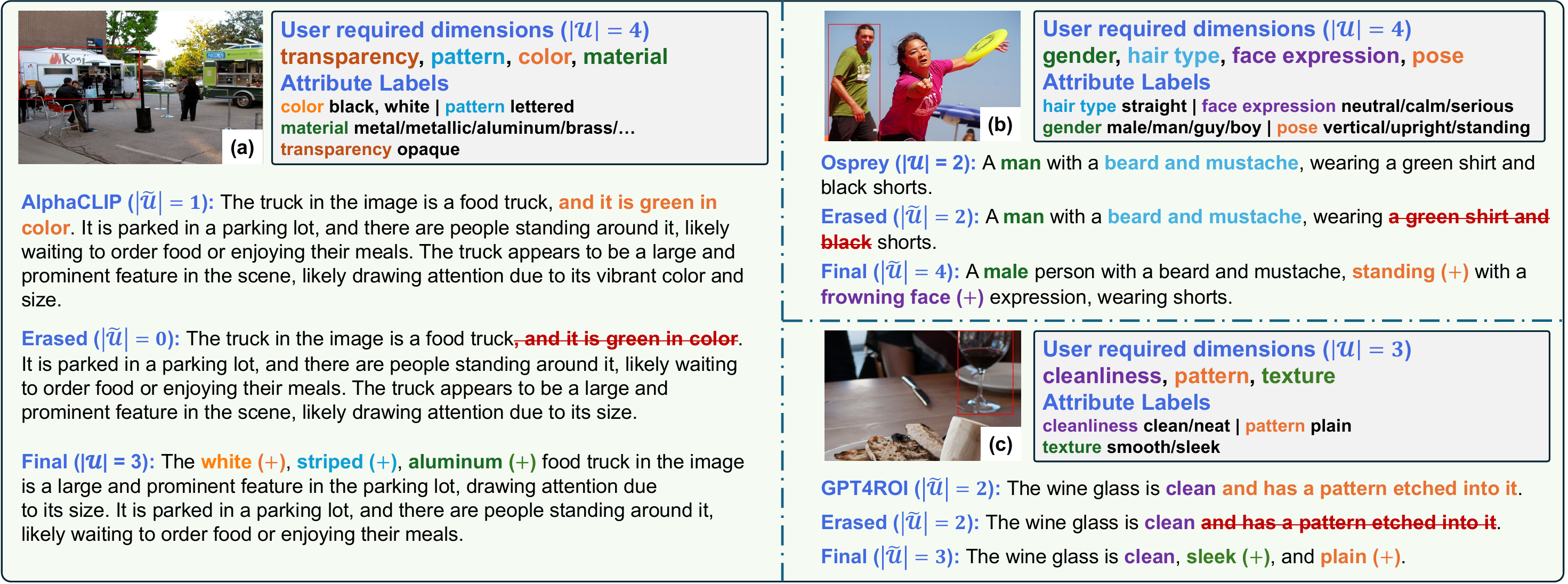}
  \caption{The visualization of the description refining process of Dimension Tailor. $|\tilde{\mathcal{U}}|$ represent the cover number of user-specified dimensions, where $\tilde{\mathcal{U}}$ = $\mathcal{U}^* \cap  \mathcal{U}$.
The bold red content highlighted by the strike-out line is the erased content. The attribute with $(+)$ is the supplemented attributes.}
  \label{fig:refine_process}
\end{figure*}

\paragraph*{\textbf{Visualization of refined results}} In Fig.\ref{fig:desp-compare}, we first compare our refined description with various MLLM-generated descriptions. The object descriptions refined by Dimension Tailor are more succinct and more in line with user requirements than the original descriptions. For instance, for the dog in Fig.\ref{fig:desp-compare} (a), our pipeline can supplement the cleanliness and texture of the dog according to the user-specified dimensions. For the horse in Fig.\ref{fig:desp-compare} (c), we can erase the redundant dimension, such as the horse's color and its well-cared state. We denote the number of user-specified dimensions correctly included in each description as $|\tilde{\mathcal{U}}|$, where $\tilde{\mathcal{U}} = \mathcal{U} \cap \mathcal{U}^*$. The high values of $|\tilde{\mathcal{U}}|$ indicate that the descriptions refined by Dimension Tailor are more comprehensive according to the user intent. We further illustrate the intermediate description of each step of the Dimension Tailor refinement process in Fig.\ref{fig:refine_process}. As exemplified in (c), Dimension Tailor has the capability to remove the erroneous attribute ``has a pattern'' and replace it with the correct attribute ``plain''. However, there are instances of failure: the BLIP may mistakenly add error attributes, such as in (a) where the pattern dimension value is incorrectly listed as "striped" instead of the correct attribute "lettered".

\begin{table}[t]
\centering
\caption{The ablation study for the role of different modules in our pipeline for controllable object description refinement. $R_m$ is the ratio of modified descriptions.}
\normalsize
\begin{tabular}{l|ccccc}
\hline \hline
MLLMs & GPT-A & mDR & mDP & mDF1 & $R_m$ \\ \hline

AlphaCLIP      & 59.3 & 70.7 & 76.8 & 70.4 & $-$      \\  \hline

\multicolumn{6}{l}{+ Dimension Erasing}  \\ \hline
                       
$\tau_e = 0$   & $62.7$ & $48.4$ & $85.1$ & $59.5$ & $79.3$      \\
$\tau_e = 0.1$ & $60.7$ & $61.1$ & $87.4$ & $69.5$ & $66.0$     \\
$\tau_e = 0.2$ & $60.2$ & $63.5$ & $87.7$ & $71.2$ & $61.1$   \\ 
$\tau_e = 0.3$ & $60.6$ & $64.8$ & $88.5$ & $72.1$ & $59.3$ \\
$\tau_e = 0.4$ & $60.1$ & $65.8$ & $88.9$ & $73.3$ & $58.4$      \\ \hline

\multicolumn{6}{l}{+ Dimension Supplementing}  \\ \hline

$\tau_c = 0$    & $59.5$ & $76.6$ & $86.3$ & $79.1$ & $76.2$  \\ 
$\tau_c = 0.3$  & $59.7$ & $72.5$ & $86.9$ & $76.9$ & $71.1$ \\  
$\tau_c = 0.6$  & $59.5$ & $68.1$ & $87.8$ & $74.3$ & $65.0$ \\ 
\hline

\multicolumn{6}{l}{+ Dimension Filtering (LLM Predicated)} \\ \hline
$\tau_c = 0$   & $60.4$ & $77.5$ & $87.5$ & $80.4$ & $76.3$\\
$\tau_c = 0.3$ & $60.2$ & $71.1$ & $87.7$ & $76.4$ & $69.8$ \\
$\tau_c = 0.6$ & $60.7$ & $66.7$ & $88.6$ & $73.5$ & $63.3$ \\ 

\hline \hline
\end{tabular}
\label{tab:ctrl-ablation}
\end{table}

\subsection{Ablation Study}

\paragraph*{\textbf{The role of dimension erasing}} The ablation study results are shown in Table \ref{tab:ctrl-ablation}. We first investigate the effect of the value of dimension erasing threshold $\tau_e$.
Utilizing the dimension erasing can significantly enhance the mDP, assisting users in filtering out redundant dimensional content efficiently. Although too many times of erasure, e.g., $\tau_e = 0$, can improve the accuracy of the description, it can also erase some of the dimensions that the user wants, resulting in a drastic drop in mDR. Therefore, we choose $\tau_e = 0.3$, which has relatively good attribute quality and controllability.

\paragraph*{\textbf{The role of dimension supplementing}} We further explore the role of dimension supplementing and the supplement threshold $\tau_c$. As shown in Table \ref{tab:ctrl-ablation}, for the controllable object descriptions, the dimension supplementing will replenish the dimensions the user desires but are missing from the original description. Increasing $\tau_c$ may enhance the attribute quality but could result in adding fewer dimensions the user desires. For AlphaCLIP + LLaVAv1.5, we choose $\tau_c = 0$, which has relatively good quality and high mDR, to refine object descriptions.

\paragraph*{\textbf{The role of dimension filtering}}

As shown in Table \ref{tab:ctrl-ablation}, after adding the dimension filtering, the quality of the description has improved. To explore the underlying reason for its effectiveness, we conduct additional analysis. The first question is about the accuracy of the attribute filtering using LLM. Does it filter out attributes that may be correct? To measure this, we count the object-attribute combination in the annotation of 1k OVAD instances. By doing so, we can obtain the ground truth object-attribute combinations: each object with attributes that can be used to describe it. After that, we calculate the intersection over union (IoU) value to measure the consistency between the ground truth combination and the LLM-predicted combination. The IoU values between the combinations predicted by LLM and the ground truth combinations of some objects are presented in Table \ref{tab:filter_iou}. The results suggest that LLM demonstrates strong performance in certain categories, such as ``car'' and ``person'', but encounters challenges in some categories, such as ``horse'' and ``bear'', whose object names are not detailed enough. The horses in the test set include not only the real animal horse but also the horse sculpture. In the ground truth, the horse sculpture's material can be stone, but LLM considers the horse as a real animal and its material cannot be stone. This results in low IoU values for these categories. Another reason for the low IoU values is that the attribute annotations in OVAD \cite{bravo-2023-CVPR-OVAD} are multiple synonyms in one label, but not all synonyms can be used to describe the object. For example, one of the attribute labels for the category ``hot dog'' of texture dimension is ``texture:soft/fluffy/furry/hairy'', but only soft can be used to describe ``hot dog''.

\begin{table}[htbp]
    \centering
    \caption{The intersection over union (IoU) values between the object-attribute combinations predicted by LLM and the ground truth object-attribute combinations of each object.}
    \normalsize
    \begin{tabular}{ccccc}
    \hline\hline
    
    \textbf{mean} & baseball bat  & horse & car & cat  \\
    $45.7$ & $54.7$ & $36.3$ & $54.4$ & $42.5$ \\ \hline
    wine glass & refrigerator & person & bear & hot dog \\
    $61.7$ &$54.4$ & $86.0$ & $36.5$ & $8.3$ \\ \hline
    tv & tie  & truck & laptop & couch  \\
    $62.6$ & $55.6$ & $49.5$ & $56.9$ & $47.6$ \\ \hline
    potted plant & chair & cake & cup & umbrella \\
    $40.5$ &$47.0$ & $86.0$ & $52.1$ & $28.7$ \\ \hline
    boat & cell phone & book & orange  & apple \\
    $47.6$ & $57.0$ &$60.7$ & $22.5$ & $55.6$ \\

    \hline\hline
    \end{tabular}
    \label{tab:filter_iou}
\end{table}

\section{Conclusion}
In this paper, we introduce Dimension Tailor, a novel training-free pipeline for refining descriptions to align with user intent dimensions. We have developed three evaluation metrics to measure the controllability of object descriptions in terms of dimensions. Through extensive experiments, we have analyzed the controllability of object descriptions produced by recent MLLMs, including the advanced commercial MLLM GPT-4o. The experimental results show that MLLMs sometimes fail to accurately capture user-specified dimensions in their generated descriptions. In contrast, our proposed method, Dimension Tailor, offers a simple yet cost-effective way to consistently improve the controllability performance of recent open-sourced MLLMs to the commercial MLLM level.

\bibliography{ref}
\bibliographystyle{IEEEtran}












\vfill

\end{document}